\documentclass[letterpaper,10pt, conference]{ieeeconf}

\IEEEoverridecommandlockouts
\usepackage{cite}
\usepackage{amsmath,amssymb,amsfonts}
\usepackage{algorithmic}
\usepackage{graphicx}
\usepackage{tabularx}
\usepackage[table]{xcolor}

\usepackage{subcaption}

\usepackage{fancyhdr}
\usepackage{lipsum}
\definecolor{lightblue}{RGB}{181, 203, 255}  

\usepackage{textcomp}
\usepackage{balance}
\usepackage{url}
\usepackage{xcolor}
\def\BibTeX{{\rm B\kern-.05em{\sc i\kern-.025em b}\kern-.08em
    T\kern-.1667em\lower.7ex\hbox{E}\kern-.125emX}}

\usepackage{float}

\usepackage{cuted}
\usepackage{capt-of}

\begin{document}

\title{\LARGE \bf Brain-Robot Interface for Exercise Mimicry\\
\thanks{Funded by EU Horizon 2020 Research and Innovation Programme (Grant ID: 101007990)}
}

\author{Carl Bettosi$^{1 \textbf{*}}$, Emilyann Nault$^{1}$, Lynne Baillie$^{1}$, Markus Garschall$^{2}$, \\
Marta Romeo$^{1}$, Beatrix Wais-Zechmann$^{2}$, Nicole Binderlehner $^{2}$, and Theodoros Georgiou$^{1}$
\thanks{$^{1}$ Heriot-Watt University, Edinburgh, UK,
        {\textbf{*}email: cb54@hw.ac.uk}
        }%
\thanks{$^{2}$ AIT Austrian Institute of Technology GmbH, Vienna, Austria
        }
}


\maketitle
\thispagestyle{empty}
\pagestyle{empty}


\begin{abstract}
For social robots to maintain long-term engagement as exercise instructors, rapport-building is essential. Motor mimicry---imitating one's physical actions---during social interaction has long been recognized as a powerful tool for fostering rapport, and it is widely used in rehabilitation exercises where patients mirror a physiotherapist or video demonstration. We developed a novel Brain-Robot Interface (BRI) that allows a social robot instructor to mimic a patient’s exercise movements in real-time, using mental commands derived from the patient’s intention. The system was evaluated in an exploratory study with 14 participants (3 physiotherapists and 11 hemiparetic patients recovering from stroke or other injuries). We found our system successfully demonstrated exercise mimicry in 12 sessions, however, accuracy varied. Participants had positive perceptions of the robot instructor, with high trust and acceptance levels, which were not affected by the introduction of BRI technology.
\end{abstract}


\section{Introduction}

Around 80\% of stroke and brain injury survivors have an upper limb impairment, limiting movement in their arm~\cite{langhorne2009motor}. This impairment often persists over the long term ($>$12 months after stroke onset) and dramatically affects their activities of daily living. Improving arm movement after injury requires repetitive exercise training~\cite{langhorne2009motor}. However, a lack of motivation~\cite{shaughnessy2006testing}, forgetfulness~\cite{aben2008memory}, and strained healthcare resources within communities often contribute to a lack of uptake in these programs. Towards addressing these challenges, social robots have been increasingly investigated over recent decades as a means to autonomously facilitate exercise~\cite{cifuentes2020social}.

In real-world settings, human instructors, be they a physical therapist or other, employ a variety of social techniques to foster stronger relationships with their patients~\cite{miciak2019framework}. One of these techniques is the real-time mimicry of a patient’s limb movements during exercises. Through this behavior, the instructor not only acts as a visual guide, moderating speed and range of motion (ROM), but, when appropriate, also motivates the patient to extend their ROM by lifting higher or pushing further. Beyond its functional role, mimicry serves a secondary purpose: it taps into an innate social behavior between humans that has shown to increase rapport and social coordination~\cite{lakin2003using}, making it a healthy strategy in cooperative situations like exercise instruction~\cite{bourgeois2008impact}.

In Human-Robot Interaction (HRI), mimicry has been shown to be a powerful tool for fostering more engaging and supportive environments~\cite{shinohara2018humanoid}. However, its application has been limited to more natural movement, such as facial expressions or the tilting of one's head during conversation~\cite{fu2024human}. Few studies have examined more deliberate mimicry for the purposes of facilitating a task. Furthermore, perceiving and responding to real-time physical movements presents unique challenges. Technologies such as computer vision and wearable sensors can provide movement data~\cite{hun2023design, boukhennoufa2022wearable}, but may struggle to differentiate intentional and non-intentional, or subtle movements, such as those by patients with severe motor impairments.

Brain-Computer Interface (BCI) technologies present a promising alternative to address this challenge. Increasingly used as an intervention in stroke and neurorehabilitation~\cite{yang2021exploring}, BCIs offer the ability to passively monitor brain activity or enable direct control of external devices. In the context of HRI, BCIs---sometimes referred to as "Brain-Robot Interfaces"~\cite{naveed2012brain}---are gaining traction. They uncover new possibilities for robots to perceive human mental states more directly which are otherwise challenging to detect with traditional means, enhancing the robot's ability to facilitate more informed social interactions~\cite{alimardani2020passive}.

In this work, we present an exploratory study with a novel BRI system designed to mimic the movements of a patient’s upper limb during rehabilitation exercises in real time. This mimicry mechanism is integrated into an overarching exercise session autonomously led by the robot. We evaluated the system through a user study conducted in an active physiotherapy setting with patients recovering from stroke and brain injury. The study aimed to:
1) assess whether our BRI could accurately facilitate real-time exercise mimicry for this patient population, and,
2) assess patient perceptions of both BRI and social robots being used together for this purpose. This paper contributes, to the authors' knowledge, the first real-world study to use BRI to facilitate human movement mimicry during an interaction task, revealing early insights and key lessons for the HRI community.

\section{Background}

\subsection{Social Robots as Exercise Instructors}

Rehabilitation exercise sessions lend themselves well to autonomous guidance due to their sequential and repetitive nature. For social robots, their anthropomorphic embodiment can allow for realistic demonstration of exercises and rich speech and gesture capabilities, facilitating more engaging interaction over the likes of virtual agents~\cite{vasco2019train}. Over the past two decades, there have been a modest number of works dedicated to social robot exercise instructors, demonstrating their effectiveness in communicating instructions~\cite{pulido2019socially}, adapting difficulty or exertion of exercise~\cite{winkle2020situ, bettosidesigning}, providing motivational progress reporting~\cite{irfan2023personalised}, and personalizing to social preferences to optimize engagement~\cite{ross2024implementation}.


For social robots to be truly useful in many real-world contexts, they must have the ability to sustain engagement over long-term interaction~\cite{fong2003survey}, as opposed to single-session interactions. For physical rehabilitation guidance, which often takes months or years, building strong rapport and trust in the instructor can be beneficial to exercise adherence~\cite{miciak2019framework}, so understanding how robots may achieve this is of importance~\cite{hancock2011meta}.

\subsection{Motor Mimicry in Social Interaction}

In human psychology, motor mimicry---the real-time mirroring of another's physical movements---is recognized as a key mechanism through which individuals strengthen trust and rapport with one another~\cite{bavelas1986show}. In cooperative, goal-based tasks such as guided exercise sessions, conscious motor mimicry can be highly engaging. Video-based exercise guides, for instance, are a common material used in the recovery of stroke patients, where they are asked to mimic the exercise in real time~\cite{mak2016effects, tang2015physio}.

In human-robot interactions, mimicry has been explored as a way to enhance trust, rapport, and perceived human-likeness. Bartkowski et al. found that a non-humanoid robot synchronizing its movements with a human's increased trust ratings~\cite{bartkowski2023sync}. Similarly, Pasternak et al. showed that mirroring head movements and facial expressions improved rapport with service robots~\cite{pasternak2021towards}. Fu et al. compared IMU sensing and computer vision for real-time facial and head movement mimicry on the iCub and Pepper robots, finding mimicry enhanced human-likeness but was limited by IMU latency~\cite{fu2024human}.

As briefly evidenced, mimicry in HRI focuses heavily on natural, passive motor movement. However, in the authors' opinion, more work is needed that investigate deliberate motor mimicry to enhance a functional task. It should be noted that various terms, e.g., \textit{mimic}, \textit{mirror}, \textit{imitate}, are used interchangeably throughout the literature to describe a process involving the robot perceiving and replicating the action(s) of the human. In this paper, we focus on \textit{mimicry} as it conveys a more deliberate, conscious act~\cite{chartrand2009human} which our system looks to facilitate. 

\subsection{The Brain-Robot Interface}

A Brain-Computer Interface (BCI) enables communication between the brain and external devices, typically via electroencephalography (EEG). By detecting neural activity through scalp electrodes, BCIs can passively monitor mental states or actively control assistive technologies~\cite{nicolas2012brain}. In stroke rehabilitation, BCIs promote brain plasticity, aiding recovery when integrated with therapy~\cite{yang2021exploring}. They allow stroke survivors to convey motion intentions non-physically, enhancing motivation and engagement in rehabilitation~\cite{coscia2019}. However, single-session interventions have limited impact on motor recovery, highlighting the need for longitudinal BCI studies.

In HRI, tasks typically require some perception of the human interaction partner to complete the feedback loop, with more complex tasks demanding a deeper understanding. Traditional perception methods like computer vision, speech, and the use of wearable sensors provide no insight into user intentions before those actions occur in the real world. Recently, Brain-Robot Interfaces (BRIs)—--relaying human brain activity information to a robot—--have been employed to achieve this deeper level of perception~\cite{ZHANG2021}. For example, Handelman et al. use a collaborative shared control strategy where a BRI was implemented to coordinate two prosthetic limbs for a bimanual self-feeding task~\cite{Handelman2022}. Additionally, BRIs are increasingly used for controlling humanoid robots~\cite{chamola2020brain} (e.g., for grasping~\cite{spataro2017}, or telepresence~\cite{saduanov2018}). Given their success, more applications are now using BRIs in healthcare for rehabilitation and cognitive assessment~\cite{joshi2024robot}. In particular, several works have started to show the positive effects that BRI systems have on stroke survivors' rehabilitation~\cite{Gandolfi2018}.

\section{System Design} \label{sec:system-design}

We developed a BRI for real-time exercise mimicry, integrated into a robot-led session where the robot guided the patient through two exercise sets.

\subsection{Hardware and Architecture}

\begin{figure}[t]
    \centering
    \includegraphics[width=\linewidth]{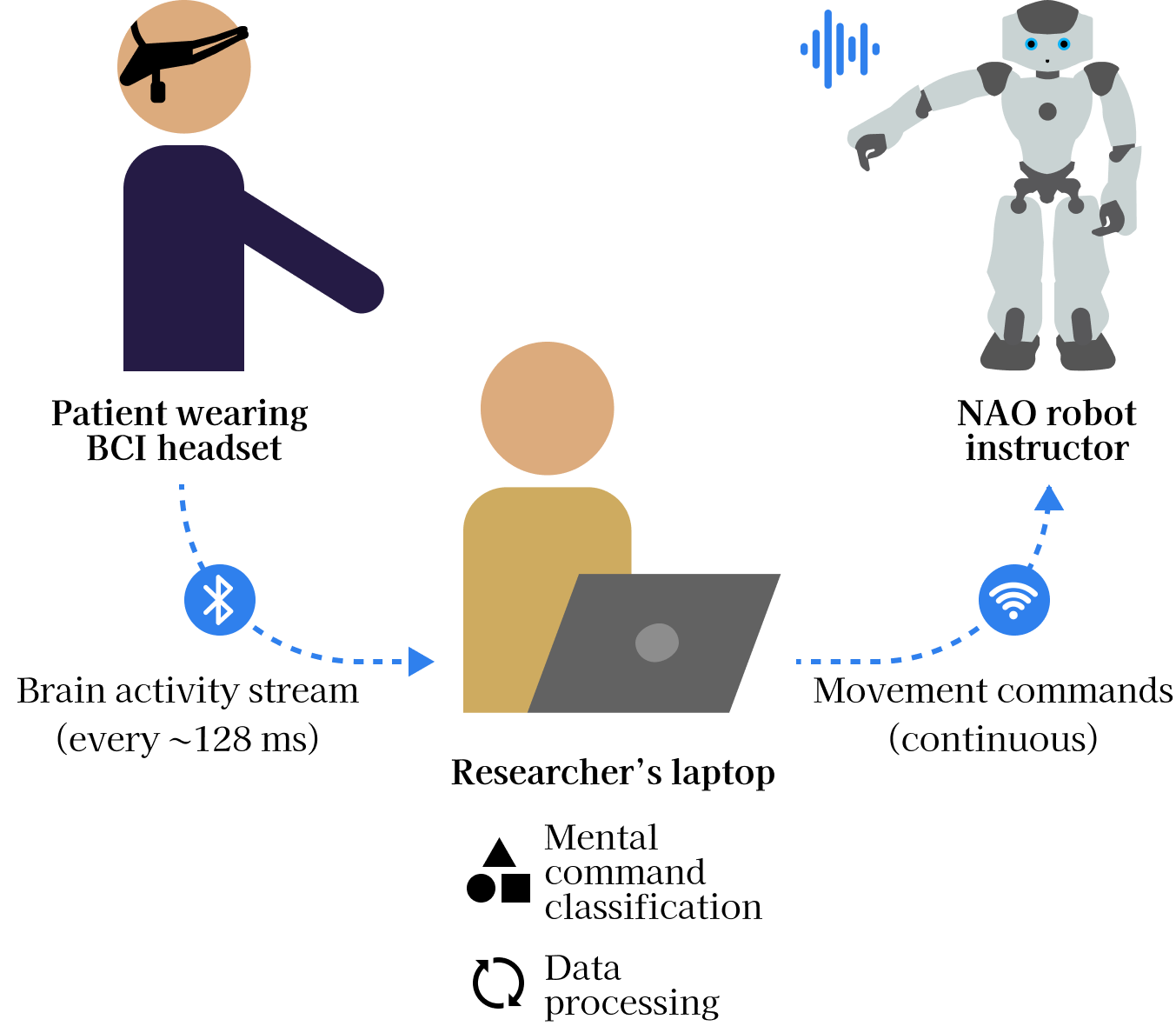}
    \caption{The system architecture complete with the BRI and robot instructor}
    \label{fig:system}
\end{figure}

We selected the Nao6 humanoid robot from Aldebaran\footnote{https://aldebaran.com/en/nao6/} as our exercise instructor due to its positive user feedback in healthcare research~\cite{nault2022investigating} and its ability to effectively mimic exercise movements with five degrees of freedom per arm. For BCI hardware, we used the Emotiv Insight 2\footnote{https://www.emotiv.com/products/}, a 5-channel EEG headset that balances accuracy, ease of setup, and cost. While more advanced systems offer higher precision, the Emotiv Insight 2 has shown good performance in previous studies~\cite{zakzouk2023brain}.

Figure \ref{fig:system} illustrates the system architecture, where the EEG headset streams data via Bluetooth to a laptop. The sensor tips use a semi-dry polymer, with a small amount of water-based gel ensuring a reliable connection through participants' hair. A Python script on the laptop manages the interaction logic and real-time data processing, converting raw EEG data into robot movements and transmitting them via WiFi to the NAO robot. Once a session begins, the robot autonomously controls the interaction, instructing, detecting, and counting exercise repetitions. 

\subsection{Real-Time Mimicry}

To achieve real-time EEG-based mimicry, we implemented a three-step process:

\label{sec:bci-training}
\subsubsection{Mental Command Training} Using EmotivBCI’s mental command feature (used successfully in previous studies on BRI control~\cite{zakzouk2023brain, joshi2024robot}), we trained the built-in EmotivBCI supervised learning classifier to distinguish task-related brain activity from neutral states (see Section \ref{sec:study-training} for details).

\subsubsection{Real-Time Processing} The EEG headset captured data at 128 samples per second per channel. The trained model classified commands as either \texttt{neutral} or \texttt{lift\_arm}. A Python script aggregated mental commands over time, transmitting the most frequent one via web sockets to the robot controller. Sampling every three commands minimized noise, reducing jitter and ensured smoother robot movement.

\subsubsection{Robot Control} \label{sec:robot_control}

The controller converted \texttt{lift\_arm} commands into incremental arm raises for the NAO robot, adjusted based on the exercise type. A \texttt{neutral} command paused movement, allowing the arm to gradually lower. A predefined threshold (45° or 90°) prevented excessive lifting. Upon reaching this limit, the robot signaled the end of the repetition.


\section{User Study} 

We recruited 22 participants: 6 physiotherapists ($\ge$5 years of experience with hemiparetic patients) and 16 hemiparetic patients (14 stroke-related, 2 from other causes). The patient group (8 females, 8 males) had a mean age of 61 years (range: 40–82). None had prior experience with social robots, though 4 had used BCI devices.

The study was conducted in Austrian German language at a community rehabilitation center in Vienna, Austria, with all responses translated into English for analysis. Ethical approval was obtained from the ethics committee of Heriot-Watt University. We first evaluated the system with the physiotherapists, but our results section in this paper largely focuses on the outcomes from the patient group.

\subsection{Protocol}

\subsubsection{Introduction and setup}

Participants signed a consent form and received an overview of the study and an introduction to the robot and BCI technology. The EEG headset was applied using primer fluid (a water-based gel) for optimal connection, with calibration supported by the EmotivBCI software. Participants sat in a stable position and wore the headset only during the training and BRI exercise sessions (Figure \ref{fig:data-stream}).

\subsubsection{Mental command training}\label{sec:study-training}

\begin{figure}[t]
    \centering
    \includegraphics[width=0.95\linewidth]{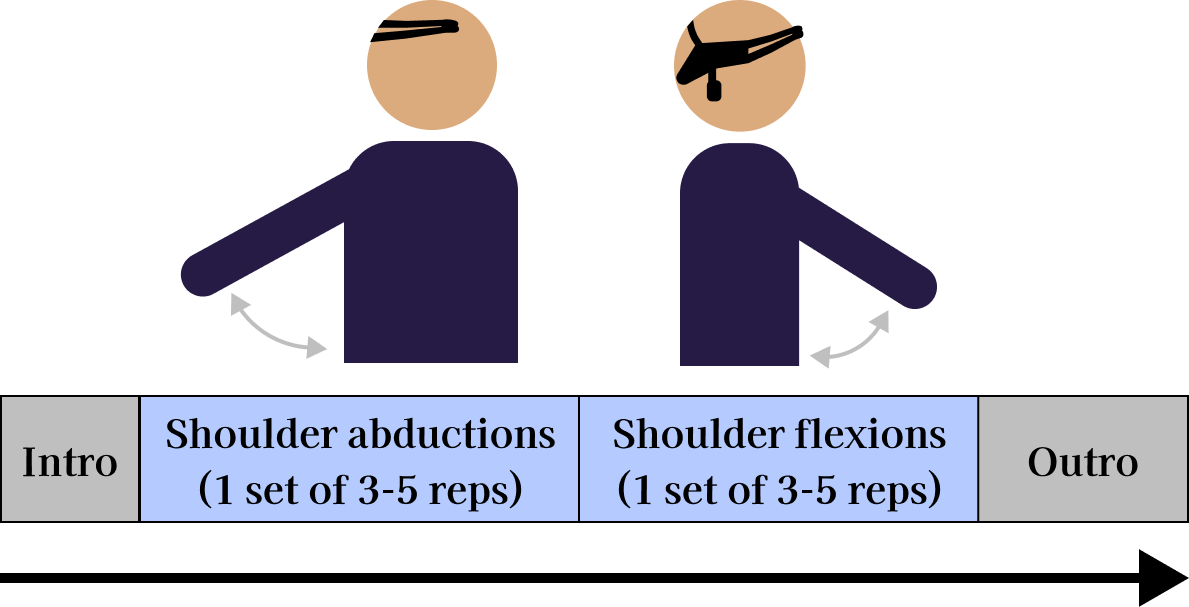}
    \caption{Structure of our robot-guided rehab session: shoulder abductions raise the arm to the side, flexion to the front. }
    \label{fig:exercise_structure}
\end{figure}

To train the system, participants completed 10–20 arm abduction repetitions (depending on their ability) while the robot demonstrated. Regardless of range of motion, the mental commands were sufficient for the EEG data to be labeled as either \texttt{neutral} or \texttt{lift\_arm}. Despite the study involving two exercises (Figure \ref{fig:exercise_structure}), early testing showed a single \texttt{lift\_arm} command was sufficient for classification, halving training requirements. Rest periods between repetitions helped train the \texttt{neutral} command. The EmotivBCI software provided real-time feedback, allowing adjustments as needed. After training, participants proceeded to the exercise sessions in a counterbalanced order.

\subsubsection{Exercise session (BRI)} \label{sec:exercise-session-1}

\begin{figure}[ht]
    \centering
      \subcaptionbox*{}[\linewidth]{%
    \includegraphics[width=\linewidth]{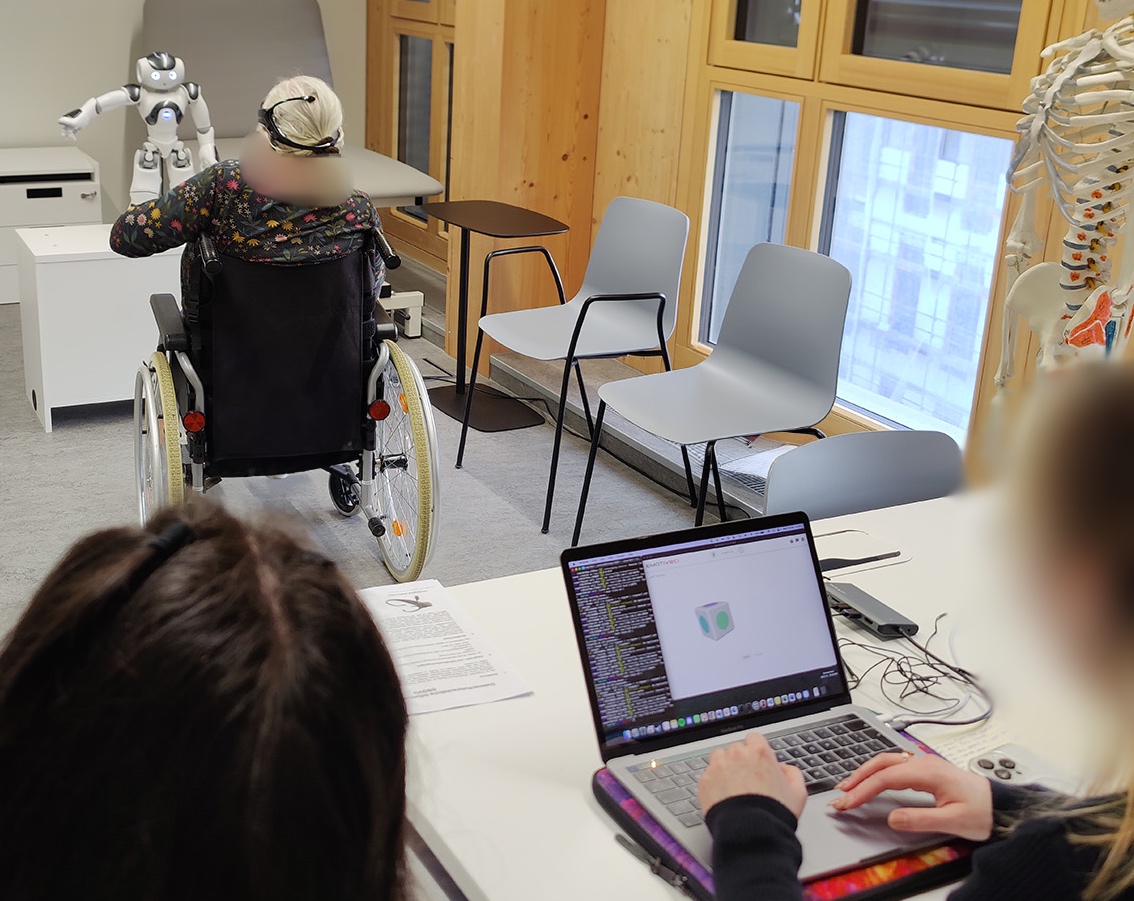}
    }

  \subcaptionbox*{}[\linewidth]{%
    \includegraphics[width=\linewidth]{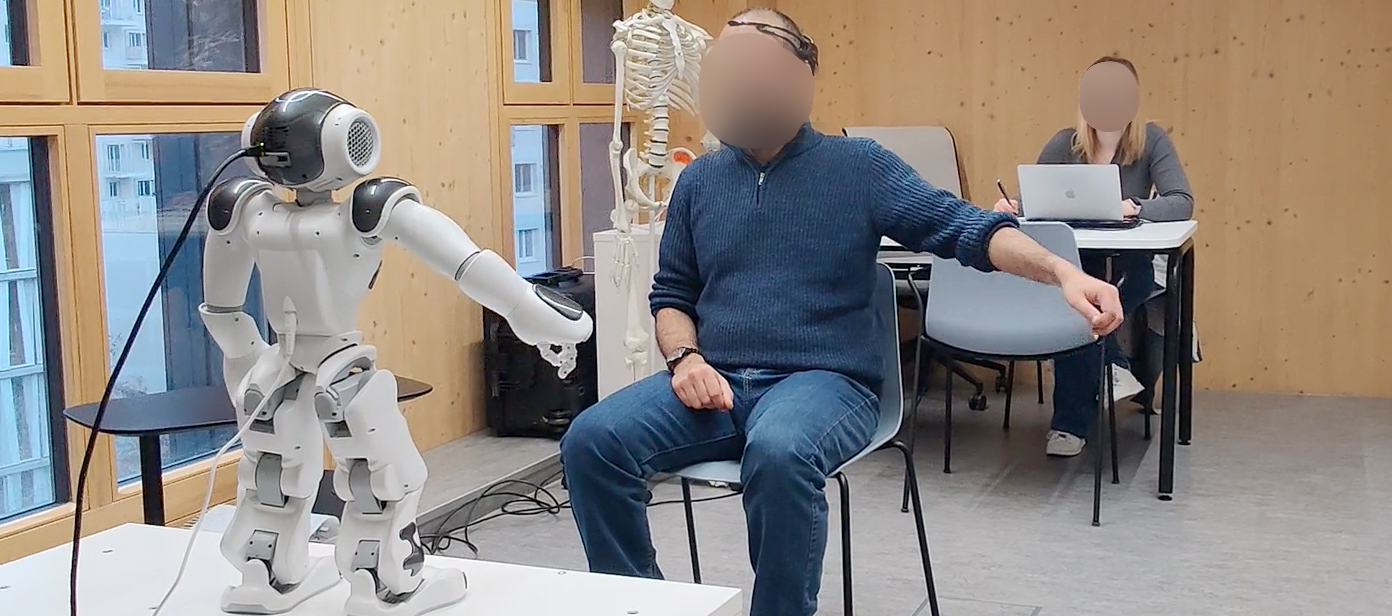}
  }
    \caption{Experiment setup during the BRI exercise session. 
    }
    \label{fig:data-stream}
\end{figure}

The participant’s ROM was recorded to set the robot’s angle threshold. The session began with the robot demonstrating two stroke recovery exercises: shoulder abduction and flexion (Figure \ref{fig:exercise_structure}). We chose these two exercises as they are considered two of the most common for stroke recovery~\cite{bettosidesigning}. To minimize fatigue, participants completed 3–5 repetitions. The robot attempted to mimic movements based on EEG commands, adapting to each participant’s pace. Two researchers monitored the session (Figure \ref{fig:data-stream}), one managing session flow and the other tracking EEG data and robot mimicry.

\subsubsection{Exercise session (non-BRI)}

Participants performed the same exercises without EEG-based control. The robot demonstrated movements but did not attempt to mimic. Researchers documented observations.

\subsubsection{Standardized questionnaires}
The Robotic Social Attributes Scale (RoSAS)~\cite{carpinella2017robotic} and the Multi-Dimensional Measure of Trust V2 (MDMT) \cite{malle2021multidimensional} were administered before and after each session to assess participants' perceptions and trust in the robot.

\subsubsection{Semi-structured interview}
To supplement questionnaire data, participants were interviewed about their experiences with robot-guided exercise and BRI technology. Key topics included perceived mimicry effectiveness, headset ease of use, and comfort during the session.

\section{Results}
Of the 22 participants, data from 8 were excluded due to technical issues—7 due to headset fitting problems (e.g., hairstyles, head shape) and 1 withdrawal due to fatigue. The final sample included 14 participants: 11 hemiparetic patients and 3 physiotherapists. Due to the exploratory nature of this study and the small sample size, we report on quantitative results but do not conduct statistical analysis, and focus on qualitative results for insight. We analyzed the qualitative data using the Constant Comparative Method (CCM) \cite{glaser1968ccm} to identify key themes.

\subsection{Exercise Mimicry}

In 12 out of 14 sessions (9 patients and all 3 physiotherapists), the robot successfully perceived enough mental commands to autonomously complete all exercise repetitions (6–10 repetitions per session), as observed by the researchers during the sessions. Participants rated the accuracy of the robot’s mimicry, that is, how well the robot was able to match their movements in real time (1=very bad, 10=very good), with an average score of 5.1 (SD=1.26). The two sessions that failed were mainly due to poor contact with the headset, meaning the researcher had to manually progress to the next phase. Notably, the training sessions for these two participants were noted to be poor or average. From the 11 patient participants that successfully completed the BRI session, they rated the headset on average 6 (SD=1.47) for ease of use (1=very difficult, 10=very easy), and 6 (SD=0.95) for comfort of the headset (1=very uncomfortable, 10=very comfortable). Interview responses also provided some additional insight into the perceived mimicry accuracy—“It’s great that the robot can imitate so well.” (P6). As an added observation, researchers also noted that mimicry was visually observed to occur in during many of the sessions where participant motor movement was clear and distinct.
 
\subsection{Social Perception} 


\subsubsection{Robotic Social Attributes Scale (RoSAS)}
The RoSAS (18-item, 1–7 scale) assessed user perceptions across warmth, competence, and discomfort \cite{Rosas1}. Warmth was rated moderately high (BRI=5.11, non-BRI=5.14), competence was moderate (BRI=4.03, non-BRI=3.74), and discomfort was low (BRI=1.70, non-BRI=1.65). Overall, patients perceived BRI and non-BRI sessions as highly similar across these dimensions.


\subsubsection{Multi-Dimensional Measure of Trust (MDMT)}
Patient trust, measured via the MDMT questionnaire, was generally high across both BRI and non-BRI sessions. Performance trust (reliability and competence) was slightly higher in non-BRI (5.85 vs. 5.63), while moral trust (ethics, transparency, benevolence) was similar (BRI: 4.98, non-BRI: 4.88). Reliability was slightly lower in BRI (5.6 vs. 5.9), but transparency was rated higher (5.8 vs. 5.23).

Participant comments supported these findings. One noted, “I trust a robot more than a human; they are more predictable” (P5). Others echoed strong confidence in the system: “My trust in the robot is pretty good anyway” (P15) and “I already have so much confidence in the system” (P14).

\subsubsection{Technology Acceptance Model} 
The TAM questionnaire~\cite{venkatesh2008tam} showed strong participant acceptance of the system, including both the robot instructor and BRI components. The perceived ease of use subscale scored an average of 1.72 out of 7 (1 = strongly agree, 10 = strongly disagree), indicating the system was easy to use. For robot anxiety, participants reported low anxiety (5.02 out of 7, where 7 = highly disagree that the robot caused anxiety). They also found the interaction enjoyable (1.94 out of 7). Overall, participants rated the system as easy to use, enjoyable, and non-anxiety-inducing.

\subsection{Key themes}
To ensure reliability, two experienced researchers independently coded 20\% of the data using the CCM mehtod. The analysis revealed three key themes: Motivation, Control, and Variation.

\subsubsection{Motivation}
Participants found the system motivating, with some suggesting improvements. One stated the robot “could serve as an inspiration” (P13), while another saw it as “a support that I can move my arm better again” (P16). Five participants cited mimicry as a motivator, with one noting, “it motivates me when someone joins in with me” (P5). The robot’s social presence also played a role, with requests for more human-like behavior, such as a natural voice or smiling when users performed well (P12, P15).

\subsubsection{Control}
Participants valued being in control rather than merely imitating the robot. One noted, “[it’s] more motivating if you can control something; just imitating [the robot] is boring” (P1), while another highlighted that controlling the robot’s movement increased engagement (P12).

\subsubsection{Variation}
Participants appreciated the robot’s ability to introduce variety, offering “more options than a therapist (P12)". They suggested it could adjust timing within sessions (P11) and be used in different settings, such as both therapy centers and homes (P16).

\section{Discussion}


In a typical scenario, we could verify the accuracy of BRI exercise mimicry by objectively comparing data to other measures such as wearable sensors or motion capture on the affected arm. However, the goal of this exploratory study was to investigate if a novel BRI could be capable of facilitating mimicry in patients with severe motor impairments. In these patients, their intention to move may not always be accompanied by clear physical actions, making traditional motion measuring methods unreliable. 

In our approach, despite some false positives (e.g., detecting a \texttt{lift\_arm} command during a \texttt{neutral} state) the robot autonomously completed all repetitions in 12 sessions (9 patients), indicating that a sufficient number of \texttt{lift\_arm} commands where sensed across sessions and this approach has potential of being successfully applied in this rehabilitation setting. In the post-interaction interview, several participants also stated that mimicry/imitation was motivating, suggesting that it worked well in those instances. However, as indicated by participant ratings, there is much room for improvement in mimicry accuracy, likely in enhancing the training process to reduce errors and improve consistency.

Overall, both physiotherapists and patients responded positively to the system, evidenced by high levels of reported trust and acceptance. Both BRI and non-BRI instructor conditions were rated low in discomfort, moderately high in warmth (RoSAS scale), and similar in perceived competence. Notably, despite the mid-level accuracy of mimicry (M=5.1, SD=1.26), BRI did not negatively impact perception, with some noting mimicry as a motivator. This suggests that improved training could enhance mimicry performance and further boost BRI's effectiveness in future studies.

\section{Limitations}

The Emotiv Insight BCI headset, chosen for its balance of accuracy, ease of use, and cost, suffered from poor signal quality in 7 participants, mainly due to small head sizes or thick hair affecting sensor contact. Frequent adjustments further distorted the fit. A more adaptable headset could improve results.

While optimal BRI function enabled reasonable movement classification with minimal samples, training was demanding due to participants' physical impairments and the single-session format. A multi-session approach would allow better data collection and integration, aligning with EEG use in stroke rehabilitation.

Finally, the small sample size (N=14) limited statistical analysis. Nonetheless, this study marks a novel integration of EEG technology with a social robot for BRI-based motor mimicry in a real-world setting.

\section{Conclusion and Future Work}
In this exploratory study, we successfully implemented and evaluated a novel Brain-Robot Interface (BRI) system for motor mimicry during upper limb rehabilitation in hemiparetic patients. To the authors' knowledge, this is the first use of BRI for motor mimicry in this context. The system demonstrated mimicry in 12 of 14 participants, though training approach, headset selection, and, subsequently, perceived mimicry accuracy are clear areas for improvement. Participants expressed high trust and acceptance of the robot, with BRI-assisted sessions showing no negative impact on perception—potentially even enhancing motivation, as suggested by qualitative feedback.

Our findings provide critical insights into the practical use of BRIs in human-robot interaction. By capturing and responding to real-time user intentions—something traditional perception technologies often fail to achieve—our system highlights the ability of BRI to enhance social interaction techniques like mimicry. This capability opens the door to more natural, responsive robot behavior, which is essential for enabling stronger rapport in long-term interactions. This has significant implications for future robots as exercise instructors, educators, and companions, where sustained, meaningful engagement is key.

\section*{Acknowledgment}
We would like to thank the Tech2People Center for Robotic Neurotherapy for their support.


\balance
\bibliographystyle{IEEEtran}
\bibliography{references}

\end{document}